\documentclass[sigconf, screen]{acmart}
\usepackage{pifont}
\usepackage{siunitx}
\acmSubmissionID{915}

\begin{document}

\title{F2Net: A Frequency-Fused Network for Ultra-High Resolution Remote Sensing Segmentation}

\author{Hengzhi Chen}
\affiliation{%
  \institution{The University of Sydney}
  \city{Sydney}
  \state{NSW}
  \country{Australia}
}

\author{Liqian Feng}
\affiliation{%
  \institution{The University of Sydney}
  \city{Sydney}
  \state{NSW}
  \country{Australia}
}

\author{Wenhua Wu}
\affiliation{%
  \institution{The University of Sydney}
  \city{Sydney}
  \state{NSW}
  \country{Australia}
}

\author{Xiaogang Zhu}
\affiliation{%
  \institution{The University of Adelaide}
  \city{Adelaide}
  \state{NSW}
  \country{Australia}
}

\author{Shawn Leo}
\affiliation{%
  \institution{Tsinghua University}
  \city{Beijing}
  \country{China}
}

\author{Kun Hu}
\affiliation{%
  \institution{Edith Cowan University}
  \state{WA}
  \country{Australia}}


\begin{abstract}
Semantic segmentation of ultra-high-resolution (UHR) remote sensing imagery is critical for applications like environmental monitoring and urban planning but faces computational and optimization challenges. Conventional methods either lose fine details through downsampling or fragment global context via patch processing. While multi-branch networks address this trade-off, they suffer from computational inefficiency and conflicting gradient dynamics during training. We propose F2Net, a frequency-aware framework that decomposes UHR images into high- and low-frequency components for specialized processing. The high-frequency branch preserves full-resolution structural details, while the low-frequency branch processes downsampled inputs through dual sub-branches capturing short- and long-range dependencies. A Hybrid-Frequency Fusion module integrates these observations, guided by two novel objectives: Cross-Frequency Alignment Loss ensures semantic consistency between frequency components, and Cross-Frequency Balance Loss regulates gradient magnitudes across branches to stabilize training. Evaluated on \textit{DeepGlobe} and \textit{Inria Aerial} benchmarks, F2Net achieves state-of-the-art performance with mIoU of 80.22 and 83.39, respectively. Our code will be publicly available.
\end{abstract}




\maketitle

\section{Introduction}

\begin{figure}[t]
    \Description{Comparison of architectures}
    \centering
    \includegraphics[width=\linewidth]{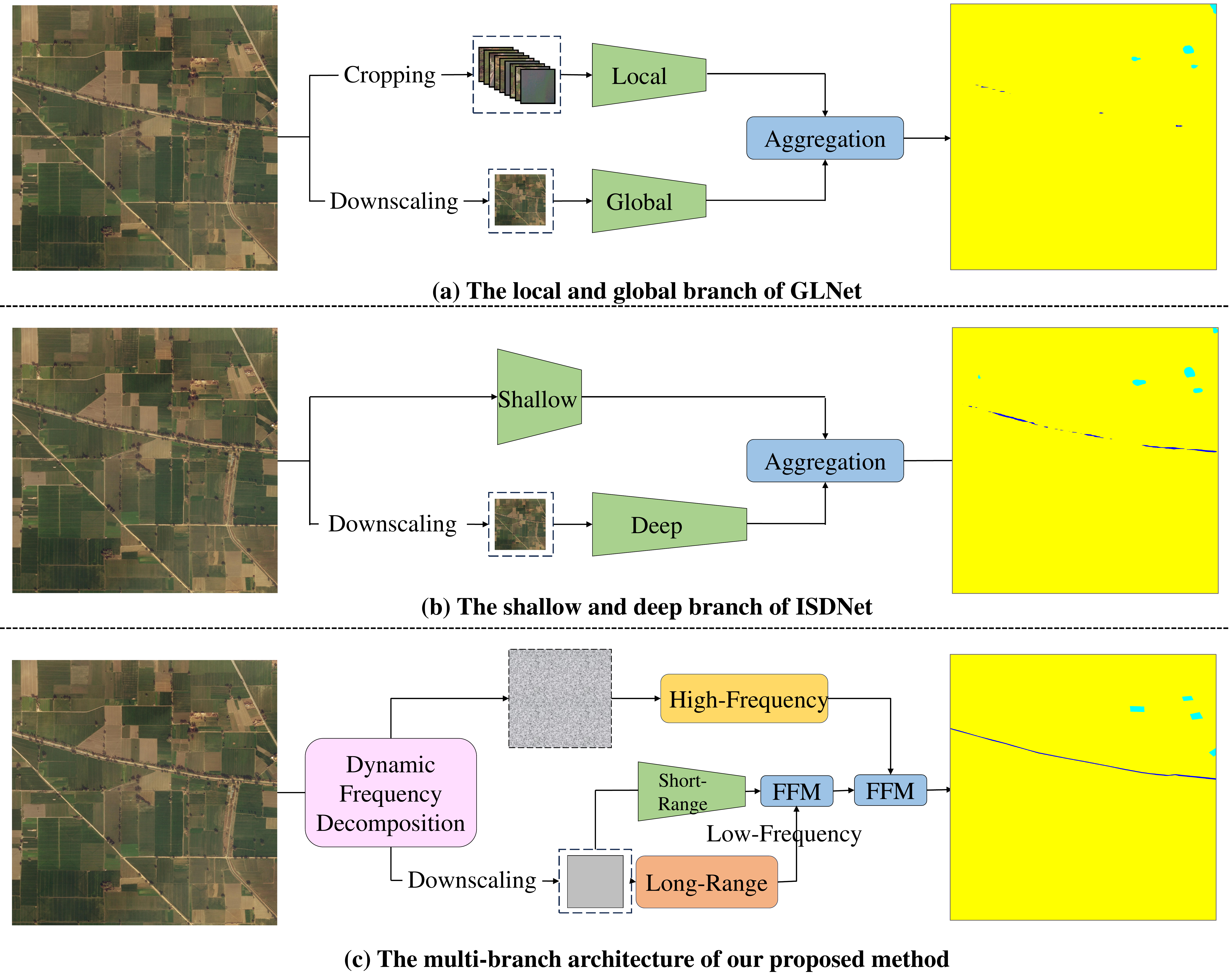}
    \caption{Comparison of multi-branch architectures for UHR RS:
(a) GLNet\cite{GLNet_2019_CVPR} uses separate branches for local and global features via patch cropping and downsampling; 
(b) ISDNet\cite{ISDNET_Guo_2022_CVPR} combines a shallow local branch and a deeper global branch; and
(c) Our F2Net applies dynamic frequency decomposition to extract high- and low-frequency observations. 
    }
    \label{compare}
\end{figure}

Semantic segmentation of remote sensing (RS) imagery is fundamental to a range of applications, including geospatial analysis~\cite{geospatial_analysis}, urban planning~\cite{urban_NEURIPS2023}, and environmental monitoring~\cite{monitor_LI2024114290}. With advances in RS technologies, ultra-high-resolution (UHR) imagery has become increasingly accessible, offering unprecedented spatial detail. While images with resolutions around $2048 \times 1080$ are conventionally considered high resolution~\cite{high_resolution_ascher1999filmmaker}, UHR typically refers to those exceeding 4K resolution~\cite{ultra_high_yu2022towards}. For segmentation tasks, UHR images provide rich spatial context, but also introduce substantial computational challenges due to their size and detail density.

Conventional semantic segmentation methods are typically designed for moderate-resolution inputs (e.g., $256 \times 256$), which are approximately 100× smaller than UHR imagery. To adapt UHR images to these networks, researchers commonly adopt one of two strategies:
downsampling~\cite{downsample_Chen_2016_CVPR,downsample2_Zhao_2018_ECCV}, which retains global contextual information but sacrifices fine-grained details essential for accurate segmentation; or
patch-based processing~\cite{patch_Strudel_2021_ICCV}, which preserves local structures but struggles to capture the global context for holistic scene understanding. To overcome the fundamental issues of using either approach in isolation, recent research has shifted toward multi-branch neural network architectures, which aim to jointly preserve fine details and incorporate broader contextual cues.

A representative example is GLNet~\cite{GLNet_2019_CVPR}, which adopts a dual-branch structure to capture global and local information jointly. It processes downsampled images and cropped patches through separate branches, enabling the model to maintain global scene awareness while preserving fine-grained local details.
ISDNet~\cite{ISDNET_Guo_2022_CVPR} builds upon this idea by eliminating patch-based processing altogether. It introduces a shallow branch that directly extracts local details from UHR inputs, which are then fused with global features obtained from downsampled images to produce a unified scene-level representation. SGNet~\cite{SGNet_wang2024toward} further generalizes this concept by proposing a surrounding context-guided branch that can be seamlessly integrated into existing segmentation models. Given a local patch, SGNet aggregates contextual cues from neighboring patches to supplement missing global information, mitigating fragmentation and enhancing boundary consistency.

While multi-branch architectures have demonstrated promising performance, they face two key challenges: computational inefficiency within individual branches and optimization conflicts during joint training.
These challenges manifest differently across designs. For instance, GLNet (Figure~\ref{compare}a) adopts a patch-cropping strategy, which requires repeated feature extraction and multiple inference passes. This results in significant computational overhead and latency for UHR images, where around 200 patches must be processed independently. Methods like ISDNet (Figure~\ref{compare}b), alleviate this by eliminating patch cropping but introduces GPU memory limitations. 
The second challenge arises during training, as parallel processing of multi-scale representations introduces competing gradient flows. This leads to unbalanced learning dynamics, causing inconsistent convergence and feature misalignment across branches. 

To address these challenges, we propose a frequency-aware network - F2Net. By decomposing UHR imagery into high-frequency and low-frequency components, F2Net comprises three frequency aware branches: a high-frequency branch for detail perservation and two low-frequency subbranches for semantic extraction.
The high-frequency branch retains full-resolution inputs and focuses on capturing fine structural details, such as edges and textures, which are critical for accurate boundary segmentation. In contrast, the low-frequency branch processes a downsampled version of the image, encoding semantically rich yet spatially redundant features. This low-frequency branch is further divided into two sub-branches: one modeling short-range dependencies and the other capturing long-range semantic context. A Hybrid-Frequency Fusion (HFF) module integrates the outputs of these branches. 
To ensure effective multi-branch training, we introduce two complementary objectives: (1) Cross-Frequency Alignment Loss (CFAL) enforces semantic consistency between high- and low-frequency representations, facilitating coherent feature integration; and (2) 
Cross-Frequency Balance Loss (CFBL) regularizes the gradient magnitudes across branches, promoting balanced learning dynamics. 
Comprehensive experiments on two widely used UHR imagery benchmarks -  \textit{DeepGlobe}~\cite{deepglobe} and \textit{Inria Aerial}~\cite{Inria_aerial_8127684} - demonstrate the effectiveness of F2Net. 

In summary, our key contributions are as follows:
\begin{itemize}
  \item A frequency-aware network - F2Net decomposes UHR RS images into high- and low-frequency components, enabling efficient and detail-preserving segmentation.
  \item We introduce two cross-frequency objectives for multi-branch leraning on UHR imagery.
  \item We achieve the state-of-the-art performance on two UHR benchmarks—\textit{DeepGlobe} and \textit{Inria Aerial}.
\end{itemize}

\section{Related Work}
\begin{figure*}[ht]
    \centering
    \Description{Overview}
    \includegraphics[width=\linewidth]{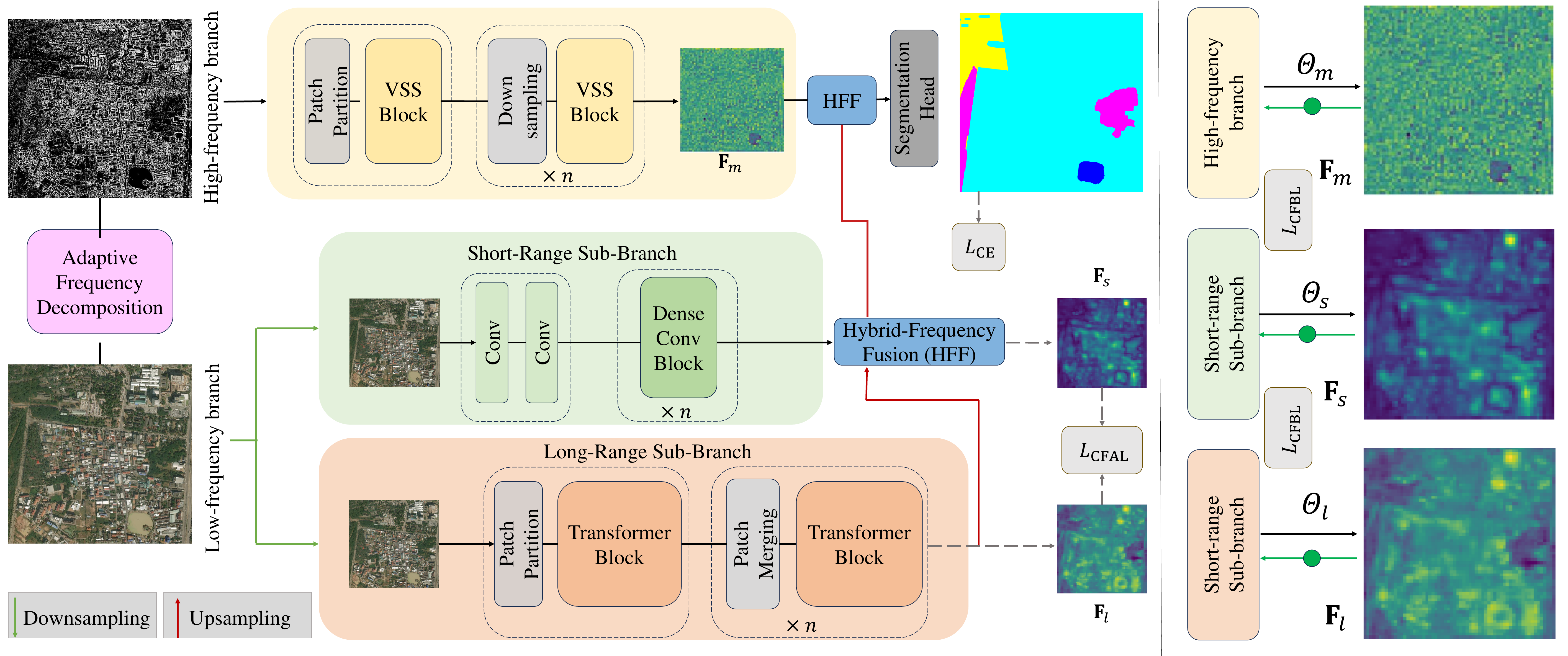}
    \caption{Overview of the proposed F2Net multi-branch architecture.
It begins with a dynamic frequency decomposition module that splits the full-resolution input into high- and low-frequency components. The high-frequency branch processes full-resolution features to preserve fine structural details. The low-frequency branch, operating on a downsampled version, consists of two sub-branches: a short-range sub-branch and a long-range sub-branch. A hybrid-frequency fusion (HFF) module integrates these complementary features.
Two auxiliary losses—Cross-Frequency Alignment Loss (CFAL) and Cross-Frequency Balance Loss (CFBL)—are introduced to enforce semantic consistency and balanced optimization across branches.
    }
    \label{main}
\end{figure*}   
\noindent\textbf{Ultra High Resolution Segmentation}.
Conventional semantic segmentation methods struggle with UHR images due to their immense size and fine-grained details. Early approaches operated UHR images with downsampling resolution or cropping patches, which compromises the details and sacrifices the global context~\cite{downsample_Chen_2016_CVPR,downsample2_Zhao_2018_ECCV}. 
Regarding capturing the long-dependency global representations, transformer-based methods are studied in the UHR segmentation field by limiting positional encodings and employing lightweight decoders to avoid higher memory usage~\cite{xie2021segformer}. 
In addition to the global information, the local details are simultaneously included by introducing hybrid CNN–Transformer architectures~\cite{zhang2022transformer}. Considering the larger scale of hybrid architecture, a surrounding context-guided module is introduced to  enhance boundary consistency by aggregating contextual cues from neighbouring patches~\cite{SGNet_wang2024toward}. 

Meanwhile, a dual-branch architecture is introduced by fusing the outcomes of downsampling and cropping to address the memory limitation~\cite{GLNet_2019_CVPR}. The global branch captures contextual information from downsampled images, and the local branch handles regional patterns on patches. 
This global-local paradigm is further extended by including multi-scale contextual features~\cite{FCTL_Li_2021_ICCV}. The proposed locality-aware contextual correlation mechanism enhanced the contour identification ability by reducing boundary artifacts but overlapping patch-based inference increased latency.
To retain the local details in capturing the global context, a combination of shallow and deep networks is presented, where the shallow branch is responsible for the spatial details globally and the deep branch focuses on semantic details locally ~\cite{ISDNET_Guo_2022_CVPR}. 
However, even with multi-branch designs, existing methods still struggle to preserve structural integrity under operations such as cropping.
In contrast, our approach retains visual integrity by decomposing the image into high- and low-frequency components, enabling the model to capture fine  details while preserving contextual consistency.


\noindent\textbf{Representation Learning in Frequency Domain}.
Converting the visual representation to the frequency domain has rapidly gained recognition in diverse remote sensing tasks owing to its ability to capture both global context and fine-grained details ~\cite{yang2024sffnet,song2023fourier,zeng2025method,ou2025focalsr}. Wavelet-based decomposition methods divide inputs into low- and high-frequency components to leverage complementary information and produce coherent predictions~\cite{li2023wavelet}. To dynamically refine these representations, adaptive frequency filtering enhances intra-class consistency while sharpening inter-class boundaries~\cite{song2023fourier}. In order to further capture the global information, Fast Fourier Convolution-based techniques contribute additional global cues, improving the detection of small objects in cluttered environments~\cite{yang2024high}. Meanwhile, discrete cosine transform and other transform-based alignment approaches facilitate robust feature matching under varying conditions, including illumination shifts or seasonal differences, extending the applicability of frequency domain representation learning across a broad range of tasks~\cite{zeng2025method}. However, approaches that incorporate multi-frequency components often face challenges related to fusion strategies and training stability - issues that remain largely unexplored in the ultra-high-resolution (UHR) remote sensing domain. 

\section{Methodology}

\subsection{Overview}
\begin{figure}[H]
\Description{Visualization of branch heatmaps}
\includegraphics[width=\linewidth]{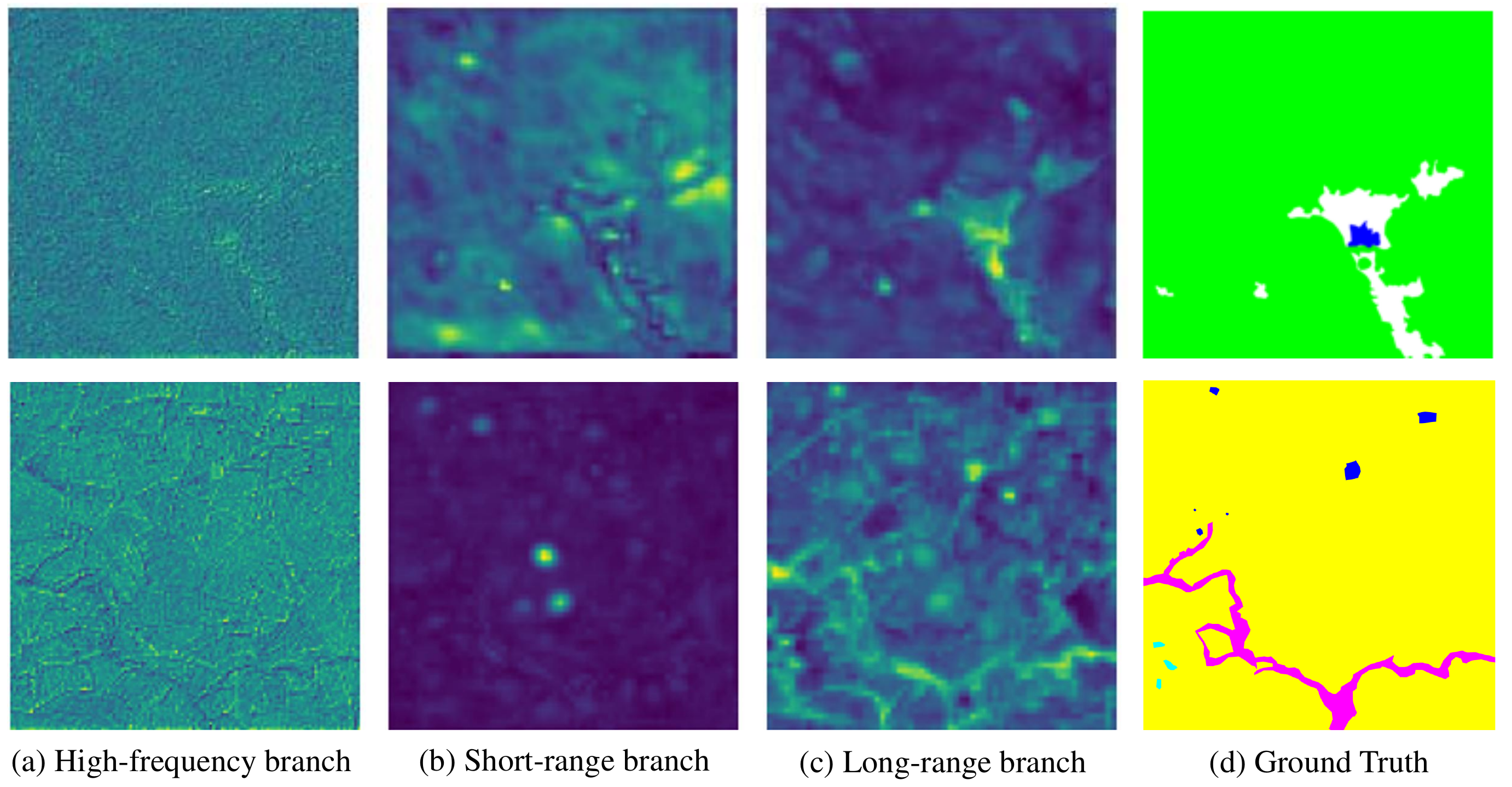}
    \caption{Visualization of branch-relevant heatmaps. (a) High-frequency branch heatmap highlights fine details; (b) Short-range branch heatmap showing significant local objects; (c) Long-range branch heatmap represents macroscopic pattern distribution; and (d) Ground truth segmentation.}
    \label{fig:heatmap}
\end{figure}

As illustrated in Figure~\ref{main}, our proposed F2Net comprises two frequency-aware branches: a high-frequency branch and a low-frequency branch. F2Net allows each branch to specialize in modeling a distinct spectral representation, simplifying the learning process and improving task focus.
The network begins with a dynamic frequency decomposition module, which separates the full-resolution input into high- and low-frequency components. 
As shown in Figure~\ref{fig:heatmap}, The high-frequency component, which retains the original resolution, is processed by the high-frequency branch to preserve fine structural details. In contrast, the low-frequency component captures semantically meaningful but spatially redundant information. This allows us to safely downsample it, significantly reducing computational cost while retaining essential context.
The low-frequency branch is further divided into two sub-branches: a short-range sub-branch that models local dependencies and a long-range sub-branch that captures global semantic context. 
A hybrid-frequency fusion (HFF) then integrates the complementary features from these branches.

Moreover, to ensure effective multi-branch training, we introduce two cross-frequency objectives: (1) Cross-Frequency Alignment Loss (CFAL) enforces semantic consistency between high- and low-frequency representations, facilitating coherent feature integration; and (2) 
Cross-Frequency Balance Loss (CFBL) regularizes the gradient magnitudes across branches, promoting balanced learning dynamics.

\subsection{Problem Formulation}
Given a UHR remote sensing image, $\mathbf{I} \in \mathbb{R}^{H \times W \times C}$, where $H$, $W$, and $C$ denote the height, width, and number of channels of the image, respectively, semantic segmentation aims to estimate a segmentation map $\mathbf{S}$, which is defined as:
    \begin{equation}
    \mathbf{S} = \{s_i^{hwl} \in \{0, 1\} \mid 1 \leq h \leq H,  1 \leq w \leq W, 1 \leq l \leq L\},
    \end{equation}
    where $s_i^{hwl}=1$ indicates that pixel at position $(h,w)$ belongs to semantic class $k$ and $s_i^{hwl}=0$ otherwise. Here, $L$ denotes the total number of semantic classes. The predicted segmentation map generated by our method is denoted as $\hat{\mathbf{S}} = \{\hat{s}_i^{hwl}\}\in [0,1]^{H\times W \times L}$.

\subsection{Adaptive Frequency Decomposition}

We begin by applying a pointwise convolution to the input image $\mathbf{I}$ for cross-channel understanding as in~\cite{magid2021dynamic}, obtaining feature maps $\mathbf{X} \in \mathbb{R}^{H \times W \times D}$:
\begin{equation}
    \mathbf{X} = \text{Conv}_{1\times1}(\mathbf{I}).
\end{equation}
A uniform kernel, typically, struggles to capture spatially varying high-frequency details, as signal frequency can change significantly across different regions of the image. To address this, we introduce a dynamic kernel formulation strategy based on channel-wise grouping. Specifically, the feature map $\mathbf{X}$ is uniformly split along the channel dimension into multiple ($N$) groups:
\begin{equation}
    \mathbf{X} = [\mathbf{X}_1, \mathbf{X}_2, \dots, \mathbf{X}_N], \quad \mathbf{X}_n \in \mathbb{R}^{H \times W \times \frac{D}{N}}.
\end{equation}
For each group $\mathbf{X}_n$, we create spatially-adaptive low-pass filters through normalized convolution:
\begin{equation}
    \mathbf{W}_n^{\text{LF}}= \text{Softmax}\left(\text{Conv}(\mathbf{X}_n)\right) \in \mathbb{R}^{H \times W\times k^2},
\end{equation}
where $k$ is the filter kernel size and the softmax operation apply along ($k^2$) dimension ensures non-negative weights that sum to one, enabling valid low-pass filtering behavior. The corresponding high-pass kernels are derived by subtracting from an identity kernel:
\begin{equation}
    \mathbf{W}_n^{\text{HF}}[h,w,:] = \mathbf{1}_{k \times k} - \mathbf{W}_n^{\text{LF}}[h,w,:],\quad \forall(h,w).
\end{equation}
These dynamic filters are then applied to their respective feature groups: 
\begin{equation}
\begin{split}
    \mathbf{X}_n^{\text{LF}} &= \mathbf{X}_n \ast \mathbf{W}_n^{\text{LF}}, \\
    \mathbf{X}_n^{\text{HF}} &= \mathbf{X}_n \ast \mathbf{W}_n^{\text{HF}},
\end{split}
\end{equation}
where $\ast$ denotes depthwise convolution. Finally, the decomposed low- and high-frequency components are reconstructed via channel-wise concatenation.
\begin{equation}
\begin{split}
    \mathbf{X}^{\text{LF}} &= [\mathbf{X}_1^{\text{LF}}, \dots, \mathbf{X}_N^{\text{LF}}] \in \mathbb{R}^{H \times W \times D}, \\
    \mathbf{X}^{\text{HF}} &= [\mathbf{X}_1^{\text{HF}}, \dots, \mathbf{X}_N^{\text{HF}}] \in \mathbb{R}^{H \times W \times D}.
\end{split}
\end{equation}

\subsection{High-Frequency Branch}

The high-frequency branch adopts a multi-stage encoder to extract hierarchical representations from the high-frequency input $\mathbf{X}^{\text{HF}}$. Since its high resolution nature, State-Space Model (SSM)~\cite{vMamba_NEURIPS2024} based visual processing is adopted for the branch as SSM has shown significant improvement in the long-sequence modeling.  
The process begins with a convolutional embedding layer, projecting the input into an initial feature space:
\begin{equation}
    \mathbf{Z}^{(0)} = \text{Conv2D}(\mathbf{\mathbf{X}^{\text{HF}}}) \in \mathbb{R}^{H_0 \times W_0 \times C_0}
\end{equation}
Feature extraction is then performed through multiple stages, each consisting of \(L\) consecutive Visual State Space (VSS) blocks. Within each block, the following operations are sequentially applied: 
1) layer normalization is used to normalize the input features:
\begin{equation}
\mathbf{Z}_{\text{norm}}^{(l)} = \text{LayerNorm}\left(\mathbf{Z}^{(l-1)}\right).
\end{equation}
Then, the SS2D module captures spatial dependencies, and its output is combined with the residual input:
\begin{equation}
\mathbf{Z}_{\text{SS2D}}^{(l)} = \mathbf{Z}^{(l-1)} + \text{SS2D}\left(\mathbf{Z}_{\text{norm}}^{(l)}\right).
\end{equation}
This is followed by a second normalization:
\begin{equation}
\mathbf{Z}_{\text{FFN}}^{(l)} = \text{LayerNorm}\left(\mathbf{Z}_{\text{SS2D}}^{(l)}\right),
\end{equation}
which is then passed through a feed-forward network (FFN). The final output of the block is computed as:
\begin{equation}
\mathbf{Z}^{(l)} = \mathbf{Z}_{\text{SS2D}}^{(l)} + \text{FFN}\left(\mathbf{Z}_{\text{FFN}}^{(l)}\right).
\end{equation}
Downsampling is applied between consecutive stages to progressively enlarge the receptive field while maintaining computational efficiency. The final output of the high-frequency branch is denoted $\mathbf{F}_m \in \mathbb{R}^{H_m \times W_m \times C_m}$.

\subsection{Low-Frequency Branch}
For comprehensive low-frequency semantic modeling, the downsampled low-frequency component $\mathbf{X}^{\text{LF}}$ is processed through two complementary subbranches. The short-range subbranch, built upon a CNN backbone - GLNet~\cite{GLNet_2019_CVPR}, focuses on capturing localized semantic features, producing an output $\mathbf{F}_s \in \mathbb{R}^{H_s \times W_s \times C_s}$.
To address the limited receptive field of CNNs, we introduce a long-range subbranch based on TinyViT~\cite{TinyVIT}, which processes the input $\mathbf{X}^{\text{LF}}$ to model long-range dependencies and generate global feature representations $\mathbf{F}_l \in \mathbb{R}^{H_l \times W_l \times C_l}$. 

\subsection{Hybrid-Frequency Fusion}
Heterogeneous features from different branches may contain redundant or misaligned channels, which can hinder effective integration. To address this, we introduce a Hybrid-Frequency Fusion (HFF) module that performs hierarchical feature aggregation.
Specifically, HFF first integrates the outputs of the two low-frequency sub-branches to capture both short- and long-range context, and then fuses this representation with the high-frequency branch output to form the final unified feature.

In detail, we first apply a channel-wise attention mechanism to each low-frequency sub-branch, enhancing informative channels while suppressing less relevant ones prior to fusion. Given $\mathbf{F}_s$ and $\mathbf{F}_l$, HFF evaluates the corresponding channel attentions $\mathbf{A}_s$ and $\mathbf{A}_l$: 
\begin{equation}
\begin{split}
    \mathbf{A}_s &= \sigma(\text{MLP}(\text{Pool}(\mathbf{F}_s)))\in\mathbb{R}^{C_s},\\
\mathbf{A}_l &= \sigma(\text{MLP}(\text{Pool}(\mathbf{F}_l)))\in\mathbb{R}^{C_l},
\end{split}
\end{equation} 
where $\sigma$ represents a sigmoid activation function, MLP denotes a multi-layer perceptron, and Pool indicates spatial average pooling.
Next, we model cross-branch relationships through: 
\begin{equation} 
\mathbf{M} = \sigma(\mathbf{A}_s \mathbf{A}_l^{\mathrm{T}}).
\end{equation}
The cross-branch attention matrix $\mathbf{M} \in\mathbb{R}^{C_s \times C_l}$ is incorporated with the original channel attention to enhance feature interaction, allowing for more effective information exchange between different branches. It is achieved via MLPs for dimension matching purposes: 
\begin{equation} 
\begin{split}
    \mathbf{\tilde{A}}_s &= \sigma(\text{MLP}(\mathbf{M}) + \mathbf{A}_s), \\
    \mathbf{\tilde{A}}_l &= \sigma(\text{MLP}(\mathbf{M}) + \mathbf{A}_l).
\end{split}
\end{equation} 
We then apply these refined attentions to their corresponding branches and fuse them together: \begin{equation} \mathbf{F}_{\text{sl}} = \text{Conv}(\mathbf{\tilde{A}}_s \cdot \mathbf{F}_s) + \text{Conv}(\mathbf{\tilde{A}}_l \cdot \mathbf{F}_l), \end{equation} where the convolution operation aligns the dimensional spaces to facilitate proper feature fusion.

The fused low-frequency feature $\mathbf{F}_{\text{sl}}$
  is subsequently combined with the high-frequency output $\mathbf{F}_m$
  via a second HFF module, yielding the final integrated representation. This comprehensive feature is then forwarded to the segmentation head to produce the final prediction.

\subsection{Cross-Frequency Loss \& Optimization}

\noindent\textbf{Cross-Frequency Alignment Loss}.
In F2Net, each frequency branch focuses on a distinct spectral component. In specific, the high-frequency branch and low-frequency branch are inherently asymmetric in both representation and semantic abstraction. Although these branches target complementary aspects of the image, they often encode different representations of the same semantic content. Without explicit alignment, this discrepancy may lead to semantic inconsistency during feature fusion and segmentation.
To enforce coherent semantic understanding across high- and low-frequency branches, we introduce a cross-frequency alignment loss (CFAL).
It encourages the branch outputs to converge at the semantic level, ensuring that the same object class is consistently represented across branches regardless of frequency bias.
We implement CFAL using a symmetric Kullback–Leibler (KL) divergence, defined as:

\begin{equation}
\mathcal{L}_{\text{CFAL}} = \frac{1}{2}\left[D_{\mathrm{KL}}(\mathbf{F}_{\text{sl}} \parallel \mathbf{F}_{\text{m}}) + D_{\mathrm{KL}}(\mathbf{F}_{\text{m}} \parallel \mathbf{F}_{\text{sl}})\right],
\end{equation}

\noindent\textbf{Cross-Frequency Balance Loss}. 
Apart from the alignment issue, the asymmetry can lead to structurally imbalanced gradient magnitudes across branches during backpropagation. 
Such imbalance can cause the optimization process to disproportionately favor one branch over the other, undermining the collaborative learning objective and resulting in inefficient frequency feature fusion.
To address this, we propose a cross-frequency balance loss (CFBL) that explicitly regularizes the gradient norms of each frequency branch, ensuring that no single branch dominates the learning dynamics.
Formally, the CFBL is defined as:

\begin{equation}
\mathcal{L}_{\text{CFBL}} = \sum_{\Theta} |G_\Theta - \bar{G}|,
\end{equation}
where $G_i(t) = ||\nabla_{\Theta} \mathcal{L}_{\text{CE}}||_2$ denotes the gradient magnitude with respect to a conventional segmentation cross-entropy loss $\mathcal{L}_\text{CE}$ for the parameters $\Theta$ of a particular branch, and $\bar{G}$ is the mean gradient magnitude across all branches. 

\noindent\textbf{Optimization}.
The overall loss 
$\mathcal{L}$ used for optimization combines the proposed cross-frequency losses: CFAL and CFBL, together with the segmentation loss $\mathcal{L}_\text{CE}$.
It is formulated as a weighted sum:
\begin{equation}
    \mathcal{L} = \lambda_1\mathcal{L}_{\text{CFAL}} + \lambda_2\mathcal{L}_{\text{CFBL}} + \lambda_3\mathcal{L}_{\text{CE}},
\end{equation}
where $\lambda_1$, $\lambda_2$, and $\lambda_3$ are hyperparameters controlling the contribution of each loss term.

\section{Experiment}

\begin{figure*}[ht]
    \centering
    \Description{deepglobe}
    \includegraphics[width=\linewidth]{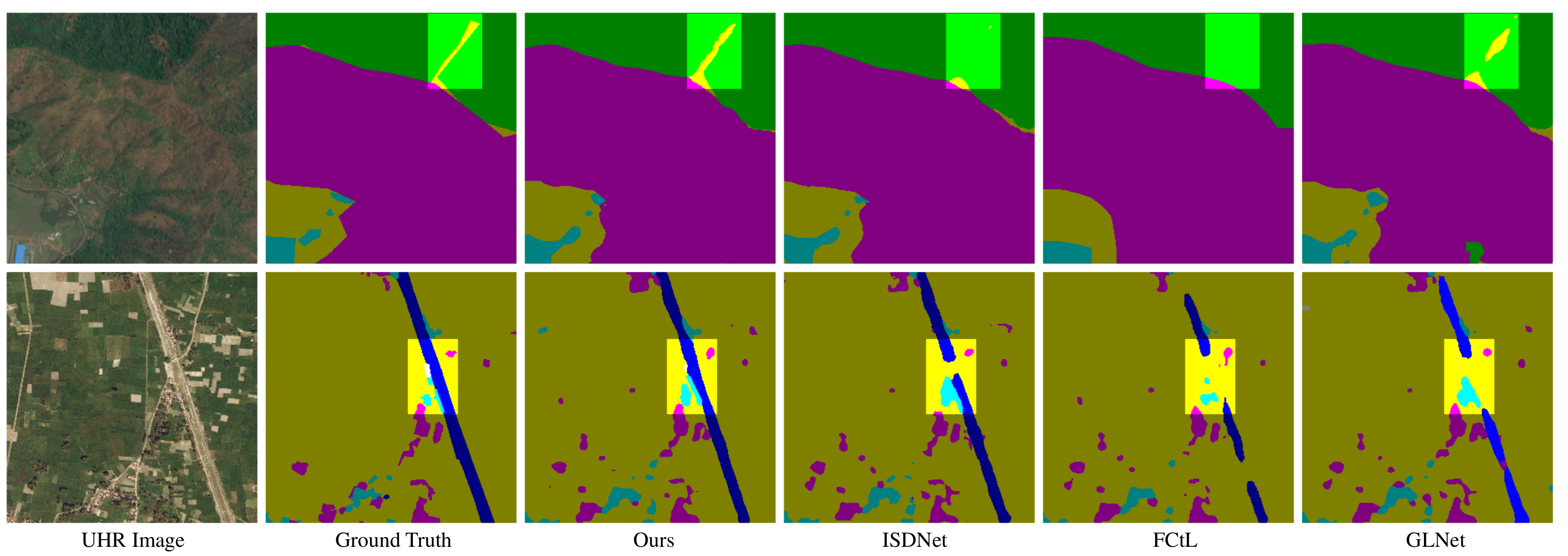}
    \caption{Illustrations of qualitative examples from the DeepGlobe dataset. Masks with different colors represent distinct semantic regions. Cyan: "\textcolor{cyan}{urban}", Yellow: "\textcolor{yellow}{agriculture}", Purple: "\textcolor{purple}{rangeland}", Green represents: "\textcolor{green}{forest}", Blue: "\textcolor{blue}{water}", White: "\colorbox{gray!25}{\textcolor{white}{barren}}", and Black: "\textcolor{black}{unknown}".
    }
    \label{visual_compare_deepglobe}
\end{figure*}

\begin{figure*}[ht]
    \centering
    \Description{Inria}
    \includegraphics[width=\linewidth]{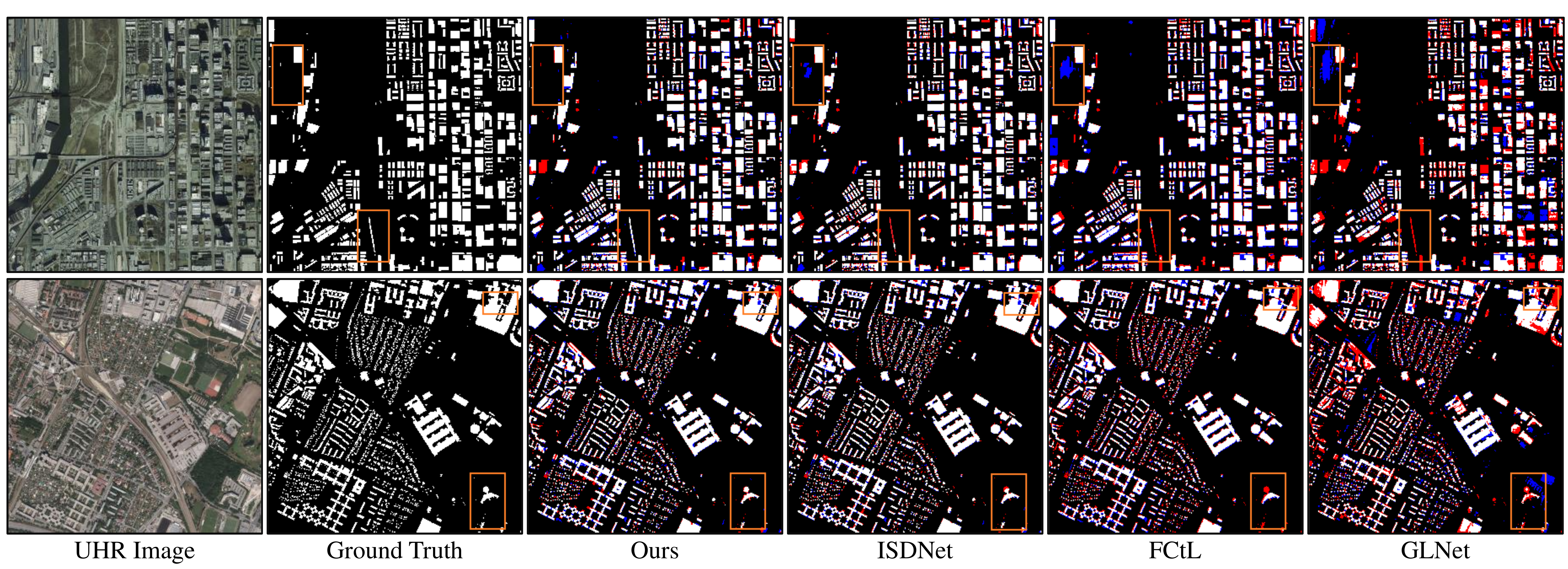}
    \caption{Illustrations of qualitative examples from the Inria Aerial dataset. White and black pixels represent building and non-building regions, respectively. For better visualization of errors, \textcolor{red}{red} pixels highlight false negatives and \textcolor{blue}{blue} pixels  indicates false positives in building detection.
    }
    \label{visual_compare_aerial}
\end{figure*}  


\subsection{Dataset}
We evaluate our method on two widely used benchmarks. 1) \textbf{DeepGlobe}~\cite{deepglobe} comprises 803 UHR satellite images, each with a resolution of $2448\times2448$ pixels. Following the protocol of~\cite{ISDNET_Guo_2022_CVPR}, the dataset was split into 455 images for training, 207 for validation, and 142 for testing. The images are densely annotated with seven landscape categories: \textit{urban}, \textit{agriculture}, \textit{rangeland}, \textit{forest}, \textit{water}, \textit{barren}, and \textit{unknown}. The unknown category is excluded from evaluation~\cite{ISDNET_Guo_2022_CVPR}.
2) \textbf{Inria Aerial}~\cite{Inria_aerial_8127684} contains 180 high-resolution aerial images ($5000\times 5000$ pixels), collected across diverse urban environments from dense metropolitan areas to alpine towns. Each image is annotated with dense binary masks distinguish buildings from non-building regions. 
Following the protocol in~\cite{FCTL_Li_2021_ICCV}, 126 images are for training, 27 for validation, and 27 for testing.

\subsection{Implementation Details}
    
The Mamba-based branch is built upon VMamba-Tiny-M2\cite{vMamba_NEURIPS2024}, with a hierarchical structure of four stages with depths [2, 2, 4, 2], and a base channel dimension 64. It leverages the selective scan mechanism introduced in Mamba2, which enables a content-adaptive receptive field that dynamically adjusts to input features. The short-range subbranch is implemented using DeepLabv3 \cite{deeplabv3+_10.1007} with a customized ResNet-18 backbone. The long-range subbranch adopts a 6-layer ViT-tiny architecture\cite{VIT_dosovits}. Both subbranches process inputs downsampled to 1/4 resolution.

All parameters are optimized using SGD with a momentum of 0.9. The initial learning rate is set to $1\times 10^{-3}$, and a polynomial decay schedule with a power of 0.9 is applied. For the Inria Aerial dataset, training is conducted for 40k iterations, while for DeepGlobe, it continues for 80k iterations. We use a batch size of 8 for all experiments, which are performed on a DGX-1 workstation equipped with Tesla V100 GPUs.

\subsection{Comparison with State-of-the-Arts}
\noindent\textbf{Quantitative Performance}.
We evaluate our F2Net against state-of-the-art methods. Performance is measured using mIOU (\%), F1 Score (\%), and Overall Accuracy (\%). The results are presented in Table~\ref{tab:segmentation_comparison_deepglobe} and Table~\ref{tab:segmentation_comparison_Aerial} for DeepGlobe and Inria Aerial datasets, respectively.
The comparison is organized into two groups: conventional segmentation methods and UHR-specific methods. As conventional models are not designed for UHR images, we adopt the evaluation protocol from prior works~\cite{ISDNET_Guo_2022_CVPR,FCTL_Li_2021_ICCV}, applying either cropping or uniform downsampling to accommodate network input constraints.
    
    
When comparing conventional segmentation methods, we observe that downsampling strategies consistently outperforms cropping. This is likely because downsampling better maintaining global structural coherence.
Overall, UHR-specific methods show clear advantages over conventional approaches, with most achieving mIOU scores above 70\%. Our proposed F2Net achieves state-of-the-art performance with nearly a 7\% improvement compared to X, and is the only method to surpass 80\% mIOU on DeepGlobe.
On the Inria Aerial dataset, UHR-specific methods again outperform adapted conventional methods. Compared to the previous best UHR method, F2Net achieves a 7\% gain in mIOU, along with improvements of 3.2\% in F1 score and 6\% in overall accuracy.

\begin{table}[t]
    \centering
    \caption{Performance comparison of different segmentation methods on \textbf{DeepGlobe}. }
    
    \begin{tabular*}{\columnwidth}{@{\extracolsep{\fill}}lcccc@{}}
    \toprule
    Method & mIoU$\uparrow$ & F1$\uparrow$ & Acc.$\uparrow$ & Memory(MB)$\downarrow$ \\
    \midrule
    \multicolumn{5}{c}{\textbf{Conventional Segmentation}} \\
    \midrule
    \multicolumn{5}{c}{\textit{Cropping}} \\
    UNet~\cite{Unet_10.1007}       & 46.53 & -  & - & \num{1741} \\
    FCN-8s~\cite{FCN_Long_2015_CVPR}     & 71.8 & 82.6  & 87.6 & \textbf{970} \\
    DeepLabv3+~\cite{deeplabv3+_10.1007}  & 63.1 & -  & - & \num{1541} \\
    \midrule
    \multicolumn{5}{c}{\textit{Downsampling}} \\
    UNet~\cite{Unet_10.1007}       & 63.5 & -  & - & \num{7627} \\
    FCN-8s~\cite{FCN_Long_2015_CVPR}     & 68.8 & 79.8  & 86.2 & \num{1984} \\
    DeepLabv3+~\cite{deeplabv3+_10.1007}  & 69.69 & -  & - & \num{3226} \\
    STDC~\cite{STDC_Fan_2021_CVPR}     & 70.30  & - & - & \num{2580} \\
    \midrule
    \multicolumn{5}{c}{\textbf{UHR Segmentation}} \\
    \midrule
    CascadePSP~\cite{cascadepsp_Cheng_2020_CVPR}  & 68.50 & 79.7  & 85.6 & \num{3236} \\
    PPN~\cite{PPN_2020}        & 71.90 & - & - & \underline{1,193} \\
    PointRend~\cite{pointrend_Kirillov_2020_CVPR}  & 71.78 & -  & - & \num{1593} \\
    MagNet~\cite{MagNet_Huynh_2021_CVPR}     & 72.96 & -  & - & \num{1559} \\
    GLNet~\cite{GLNet_2019_CVPR}       & 71.60 & 83.2  & 88.0 & \num{1973} \\
    FCtL~\cite{FCTL_Li_2021_ICCV}   & 73.5 & 83.8 & 88.3 & \num{4450} \\
    ISDNet~\cite{ISDNET_Guo_2022_CVPR}    & 73.30 & \underline{85.18} & \underline{90.17} & \num{1948} \\
    SGNet~\cite{SGNet_wang2024toward}    & \underline{75.44} & - & - & \num{2187} \\
    Ours      & \textbf{80.22} & \textbf{87.09} & \textbf{96.35} & \num{2767} \\
    \bottomrule
    \end{tabular*}
    \label{tab:segmentation_comparison_deepglobe}
\end{table}

\begin{table}[t]
    \centering
    \caption{Performance comparison of different segmentation methods on \textbf{Inria Aerial}.}
    
    \begin{tabular*}{\columnwidth}{@{\extracolsep{\fill}}lcccc@{}}
    \toprule
    Method & mIoU$\uparrow$ & F1$\uparrow$ & Acc.$\uparrow$ & Memory(MB)$\downarrow$ \\
    \midrule
    \multicolumn{5}{c}{\textbf{Conventional Segmentation}} \\
    \midrule
    FCN-8s~\cite{FCN_Long_2015_CVPR}     & 69.10 & 81.70 & 93.60 & \textbf{2,447} \\
    DeepLabv3+~\cite{deeplabv3+_10.1007}  & 55.90 &  —    &  —    & 5,122 \\
    STDC~\cite{STDC_Fan_2021_CVPR}       & 72.44 &  —    &  —    & 7,410 \\
    \midrule
    \multicolumn{5}{c}{\textbf{UHR Segmentation}} \\
    \midrule
    GLNet~\cite{GLNet_2019_CVPR}         & 71.20 &  —    &  —    & \underline{2,663} \\
    CascadePSP~\cite{cascadepsp_Cheng_2020_CVPR} & 69.40 & 81.80 & 93.20 & 3,236 \\
    FCtL~\cite{FCTL_Li_2021_ICCV}        & 73.70 & 84.10 & 94.60 & 4,332 \\
    ISDNet~\cite{ISDNET_Guo_2022_CVPR}   & 74.23 & \underline{86.35} & \underline{94.85} & 4,680 \\
    SGNet~\cite{SGNet_wang2024toward}    & \underline{81.21} &  —   &  —   &  —   \\
    Ours                                  & \textbf{83.39} & \textbf{91.19} & \textbf{98.10} & 5,534 \\
    \bottomrule
    \end{tabular*}
    \label{tab:segmentation_comparison_Aerial}
\end{table}

    \noindent\textbf{Qualitative performance}.
    We present qualitative results in Fig.~\ref{visual_compare_deepglobe} and Fig.~\ref{visual_compare_aerial}, with rectangular boxes highlighting critical regions for comparative analysis.
    In the DeepGlobe dataset (Fig.~\ref{visual_compare_deepglobe}), the top row example reveals a challenging transition zone in the upper right corner where agricultural land interfaces with forest. Due to the spectral and textural similarities between these land cover types, existing state-of-the-art methods fail to establish accurate class boundaries, erroneously classifying the entire region as forest. Our proposed approach, however, successfully delineates the agricultural areas with precise boundary definition and morphological fidelity, demonstrating superior discriminative capability in areas of high inter-class similarity. In the second row example, our method achieves continuous, topologically correct segmentation of the central watercourse, while competing approaches produce fragmented results with disconnected segments. This improvement can be attributed to our framework's ability to process non-cropped, full-resolution imagery, enabling our multi-branch architecture to leverage complementary features across multiple scales and capture spatial context that would otherwise be lost through conventional patch-based processing.
    
    In the Inria Aerial dataset (Fig.~\ref{visual_compare_aerial}), examining the first row's highlighted areas in the upper left corner, we can see that other methods incorrectly identify roads as buildings. While these areas might have textures locally similar to buildings, they clearly don't belong to building regions when considered from a global perspective. Our method correctly identifies these as non-building areas. Additionally, our approach demonstrates excellent recognition capabilities for extremely small local features, such as the thin, elongated building in the highlighted area at the bottom of the image.
    
    The second row of results showcases our method's superiority in extracting high-frequency details. In the upper right corner, buildings with complex structures and textures are shown, and our method produces more complete and detailed segmentation results compared to the alternatives.

\subsection{Ablation Studies}
To comprehensively evaluate our design, we conduct ablation studies on the DeepGlobe dataset, examining the contribution of each branch, the effect of input resolution, the impact of proposed loss functions, and the efficiency of the frequency fusion module.

\noindent\textbf{Effectiveness of frequency-aware branches}.
We first evaluate the contribution of each component in our proposed F2Net.
As shown in Table~\ref{tab:ablation_branches}, the high-frequency branch alone achieves 72.5\% mIoU, highlighting its role in capturing fine-grained boundary details.
Adding the short-range sub-branch further improves performance by 4.0, contributing complementary spatial semantic information.
The full three-branch configuration achieves 80.22\% mIoU, significantly outperforming the average of individual branches (71.9\%), indicating a strong synergistic effect.
These results confirm the effectiveness of our frequency-aware design in integrating boundary precision, local semantics, and global context.

\begin{table}[t]
\centering
\caption{Performance impact of branch combinations on DeepGlobe.}
\label{tab:ablation_branches}
\begin{tabular*}{0.95\columnwidth}{@{\extracolsep{\fill}}ccc|cc@{}}
\toprule
High-frequency & Short-range & Long-range & mIoU$\uparrow$ \\
\midrule
\ding{51} &  &  & 72.5 \\
& \ding{51} & & 71.3 \\
& & \ding{51} & 71.9 \\
& \ding{51} & \ding{51} & 73.7 \\
\ding{51} &  & \ding{51} & 76.5 \\
\ding{51} & \ding{51} &  & \underline{76.7} \\
\ding{51} & \ding{51} & \ding{51} & \textbf{80.22} \\
\bottomrule
\end{tabular*}
\end{table}

\noindent\textbf{Impact of input resolution}. 
Table~\ref{tab:ablation_resolution} shows that higher-resolution inputs consistently improve segmentation performance, particularly for the detail-aware branch.
Maintaining full resolution in this branch is critical, as performance drops by 5.1\% and 7.4\% when input is reduced to half and quarter resolution, respectively.
In contrast, downsampling the local and global branches from 1/4 to 1/8 resolution leads to only a 1.7\% drop in mIoU, while reducing memory consumption by 23\%.

Due to the memory limitations of widely-used 16GB GPUs, the input resolution for low-frequency branches cannot exceed half of the original image size during \textit{training}.
These results validate our frequency-aware design principle: high-frequency components require full-resolution processing to preserve fine details, while low-frequency semantic features can be efficiently modeled at lower resolutions—enabling a better trade-off between performance and computational cost.

\begin{table}[t]
\centering
\caption{Resolution impacts on accuracy–memory trade‑offs on DeepGlobe.}
\label{tab:ablation_resolution}
\begin{tabular*}{\columnwidth}{@{\extracolsep{\fill}}ccccc@{}}
\toprule
High‑freq. & Short‑range & Long‑range & mIoU$\uparrow$ & Mem (MB)$\downarrow$ \\
\midrule
Full & 1/4 & 1/4 & 78.2 & 2,767 \\
Full & 1/8 & 1/8 & 76.5 & 2,133 \\
1/2  & 1/4 & 1/4 & 75.1 & 2,789 \\
1/4  & 1/4 & 1/4 & 72.8 & 2,283 \\
1/4  & 1/2 & 1/2 &  —   &   —   \\
\bottomrule
\end{tabular*}
\end{table}

\begin{figure}[t]
    \centering
    \includegraphics[width=\linewidth]{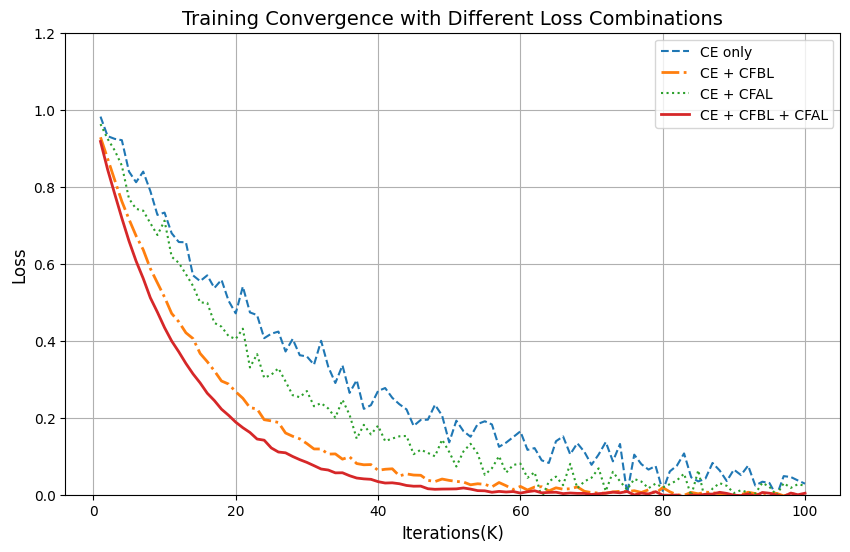}
    \caption{Training convergence with different loss combination strategies. CFBL accelerates convergence and stabilizes training.}
    \label{fig:loss_convergence}
\end{figure}

\noindent\textbf{Effectiveness of loss functions}.
We evaluate the contributions of the proposed loss components, as summarized in Table~\ref{tab:ablation_loss} and visualized in Figure~\ref{fig:loss_convergence}. The baseline employing only the standard Cross-Entropy (CE) loss achieves an mIoU of 77.35\%. Incorporating the Cross-Frequency Balance Loss (CFBL) enhances performance by 1.45\%, reaching 78.80\%, highlighting its effectiveness in balancing gradient magnitudes across branches to ensure stable and efficient training. Similarly, the integration of Cross-Frequency Alignment Loss (CFAL) alone improves the mIoU by 1.85\% to 79.20\%, demonstrating its critical role in maintaining semantic consistency between frequency-specific features.
The simultaneous application of CFBL and CFAL, together with CE, achieves the highest performance, attaining an mIoU of 80.22\%. This combined approach surpasses the CE-only baseline by 2.87\%, confirming the complementary nature and synergistic effectiveness of these novel losses.

The convergence curves depicted in Figure~\ref{fig:loss_convergence} further substantiate these findings. Particularly, incorporating CFBL significantly accelerates initial convergence and stabilizes training, as evidenced by a reduction in loss variance from ±0.08 (CE-only) to ±0.02 (CE+CFBL). Thus, our proposed loss components not only improve accuracy but also enhance the robustness and efficiency of model training.

\begin{table}[t]
\centering
\caption{Ablation on loss functions.}
\label{tab:ablation_loss}
\begin{tabular*}{0.95\linewidth}{@{\extracolsep{\fill}} ccc|c }
\toprule
$\mathcal{L}_{\text{CE}}$ & $\mathcal{L}_{\text{CFBL}}$ & $\mathcal{L}_{\text{CFAL}}$ & mIoU$\uparrow$ \\
\midrule
\ding{51} & & & 77.35 \\
\ding{51} & \ding{51} & & 78.80 \\
\ding{51} & & \ding{51} & 79.20 \\
\ding{51} & \ding{51} & \ding{51} & \textbf{80.22} \\
\bottomrule
\end{tabular*}
\end{table}

\noindent\textbf{Hybrid-Frequency Fusion Analysis}.
To validate the effectiveness of our hybrid-frequency fusion (HFF), we compare it against several common fusion strategies including feature concatenation (Concat), simple addition (Add), Atrous Spatial Pyramid Pooling (ASPP), and cross-attention. As shown in Table~\ref{tab:ablation_fusion}, our proposed HFF achieves the highest segmentation accuracy (80.22\% mIoU) with moderate computational complexity. The simple addition has the lowest complexity but provides limited accuracy (74.6\%). Concatenation slightly improves performance to 76.1\%, with minimal computational overhead. ASPP achieves a moderate mIoU of 77.3\% but incurs higher computational costs. Cross-attention, despite its significantly higher computational demand (measured in GFLOPs), results in inferior segmentation accuracy (72.4\%), highlighting the effectiveness and efficiency of our hybrid-frequency fusion approach.

\begin{table}[t]
\centering
\caption{Ablation on feature fusion strategies.}
\label{tab:ablation_fusion}
\begin{tabular*}{\columnwidth}{@{\extracolsep{\fill}}lcc@{}}
\toprule
Fusion Method  & mIoU$\uparrow$ & GFLOPs$\downarrow$ \\
\midrule
Concat         & 76.10 & 48.7 \\
Add            & 74.60 & 45.2 \\
ASPP           & 77.30 & 61.5 \\
Cross‑attention & 72.40 & 93.8 \\
Our HFF        & \textbf{80.22} & 60.8 \\
\bottomrule
\end{tabular*}
\end{table}

\noindent\textbf{Frequency Decomposition Analysis}.
Finally, we analyze the impact of our Adaptive Frequency Decomposition (AFD) on the high-frequency branch. Table~\ref{tab:ablation_gpdm} shows that AFD significantly enhances fine detail preservation while remaining high inference speed compared to direct processing of full-resolution images.

\begin{table}[t]
\centering
\caption{Ablation on frequency decomposition strategies.}
\label{tab:ablation_gpdm}
\begin{tabular*}{\columnwidth}{@{\extracolsep{\fill}}lcc@{}}
\toprule
Method & mIoU$\uparrow$ & FPS$\uparrow$ \\
\midrule
Raw Input          & 74.20 & 11.3 \\
Gaussian filter    & 76.00 & 12.1 \\
Laplacian Pyramid  & 78.10 &  9.8 \\
AFD (Ours)         & \textbf{80.22} & 15.4 \\
\bottomrule
\end{tabular*}
\end{table}

\section{Limitation}

The proposed approach lies in the fixed two-level frequency decomposition paradigm, which separates features into solely high- and low-frequency components. While this coarse split is effective for preserving details and contextual semantics, it lacks the adaptive ability to handle tasks that require more nuanced multi-scale representations. A hierarchical decomposition, such as wavelet-based or multi-band strategies, could potentially offer adequate frequency control and richer feature modeling. Moreover, despite memory-saving strategies for low-frequency branches, the overall architecture still incurs additional computation due to its multi-branch nature, which could hinder scalability for low-power applications. We leave these open questions for future exploration.

\section{Conclusion}

We propose F2Net, a frequency-aware framework for efficient semantic segmentation of UHR remote sensing images. By decomposing inputs into high- and low-frequency components, F2Net processes structural details and contextual semantics in parallel through specialized branches. The Hybrid-Frequency Fusion module integrates multi-scale features while our cross-frequency objectives stabilize training. Experiments on two widely-used benchmarks -  \textit{DeepGlobe} and \textit{Inria Aerial} - show state-of-the-art results regarding mIoU of 80.22 and 83.39, respectively. This work paves the way for efficient, high-precision RS image analysis.

\newpage


\bibliographystyle{ACM-Reference-Format}
\bibliography{sample-base}

\appendix

\end{document}